\documentclass[10pt,twocolumn,letterpaper]{article}

\usepackage{iccv}
\usepackage{times}
\usepackage{epsfig}
\usepackage{graphicx}
\usepackage{amsmath}
\usepackage{amssymb}
\usepackage{booktabs}
\usepackage{enumitem}
\usepackage{multirow}
\usepackage{color,xcolor}

\definecolor{dg}{HTML}{32CD32}
\definecolor{wjh}{rgb}{1,0,0}

\usepackage[breaklinks=true,bookmarks=false]{hyperref}

\iccvfinalcopy 

\def\iccvPaperID{2650} 

\ificcvfinal\fi

\begin{document}

\title{Ord2Seq: Regarding Ordinal Regression as Label Sequence Prediction}
\author{\textbf{Jinhong Wang}\textsuperscript{\rm 1}\thanks{Equal contribution.}\;, \textbf{Yi Cheng}\textsuperscript{\rm 1$*$}, \textbf{Jintai Chen}\textsuperscript{\rm 1$\dagger$}, \textbf{TingTing Chen}\textsuperscript{\rm 1}, \textbf{Danny Chen}\textsuperscript{\rm 2}, \textbf{Jian Wu}\textsuperscript{\rm 1}\thanks{Corresponding authors.} \vspace{0.2cm} \\
    \textsuperscript{\rm 1}Zhejiang University
    \quad \quad \textsuperscript{\rm 2}University of Notre Dame\\
 {\tt \small \{wangjinhong, chengy1, wujian2000\}@zju.edu.cn} \\
{\tt \small \{jtchen721, ttchen0603\}@gmail.com}
{\tt \small \quad dchen@nd.edu}}    \vspace{-0.2cm}

\maketitle
\ificcvfinal\thispagestyle{empty}\fi

\begin{abstract}
   Ordinal regression refers to classifying object instances into ordinal categories. It has been widely studied in many scenarios, such as medical disease grading and movie rating. Known methods focused only on learning inter-class ordinal relationships, but still incur limitations in distinguishing adjacent categories thus far. In this paper, we propose a simple sequence prediction framework for ordinal regression called Ord2Seq, which, for the first time, transforms each ordinal category label into a special label sequence and thus regards an ordinal regression task as a sequence prediction process. In this way, we decompose an ordinal regression task into a series of recursive binary classification steps, so as to subtly distinguish adjacent categories. Comprehensive experiments show the effectiveness of distinguishing adjacent categories for performance improvement and our new approach exceeds state-of-the-art performances in four different scenarios. Codes are available at \url{https://github.com/wjh892521292/Ord2Seq}.
\end{abstract}

\section{Introduction}
\label{sec:intro}

Ordinal regression, \textit{a.k.a.} ordinal classification, aims to classify object instances into ordinal categories. Since such categories follow a natural order, an ordinal regression task is typically treated as a classification problem with a few regression properties. Common applications are medical image grading~\cite{chylack1993lens, liu2018ordinal} (e.g., cataract can be graded from 0 to 6, representing \textit{normal} to \textit{severe} states), age estimation~\cite{niu2016ordinal, pan2018mean, li2019bridgenet, wen2020adaptive}, historical image dating~\cite{palermo2012dating, martin2014dating}, and image aesthetic grading~\cite{kong2016photo, lee2019image, pan2019image}.

Unlike general classification tasks, it is challenging to distinguish the adjacent categories due to their confusing data patterns and blurred boundaries in ordinal regression tasks.
Previous works often highlighted the ordering relations by introducing $K$-rank algorithms~\cite{frank2001simple, li2006ordinal, niu2016ordinal, chen2017using}, ordinal distribution constraint assumptions~\cite{lim2019order, liu2019probabilistic, lee2020deep, li2021learning}, soft labels~\cite{fu2018deep, diaz2019soft}, or multi-instance comparing approaches~\cite{liu2017deep, liu2018constrained, li2021learning, shin2022moving}. However, these methods failed to specifically tackle the ``adjacent categories distinction'' and hinder the model performances. 

\begin{figure}[t]
\centering
\includegraphics[width=0.42\textwidth]{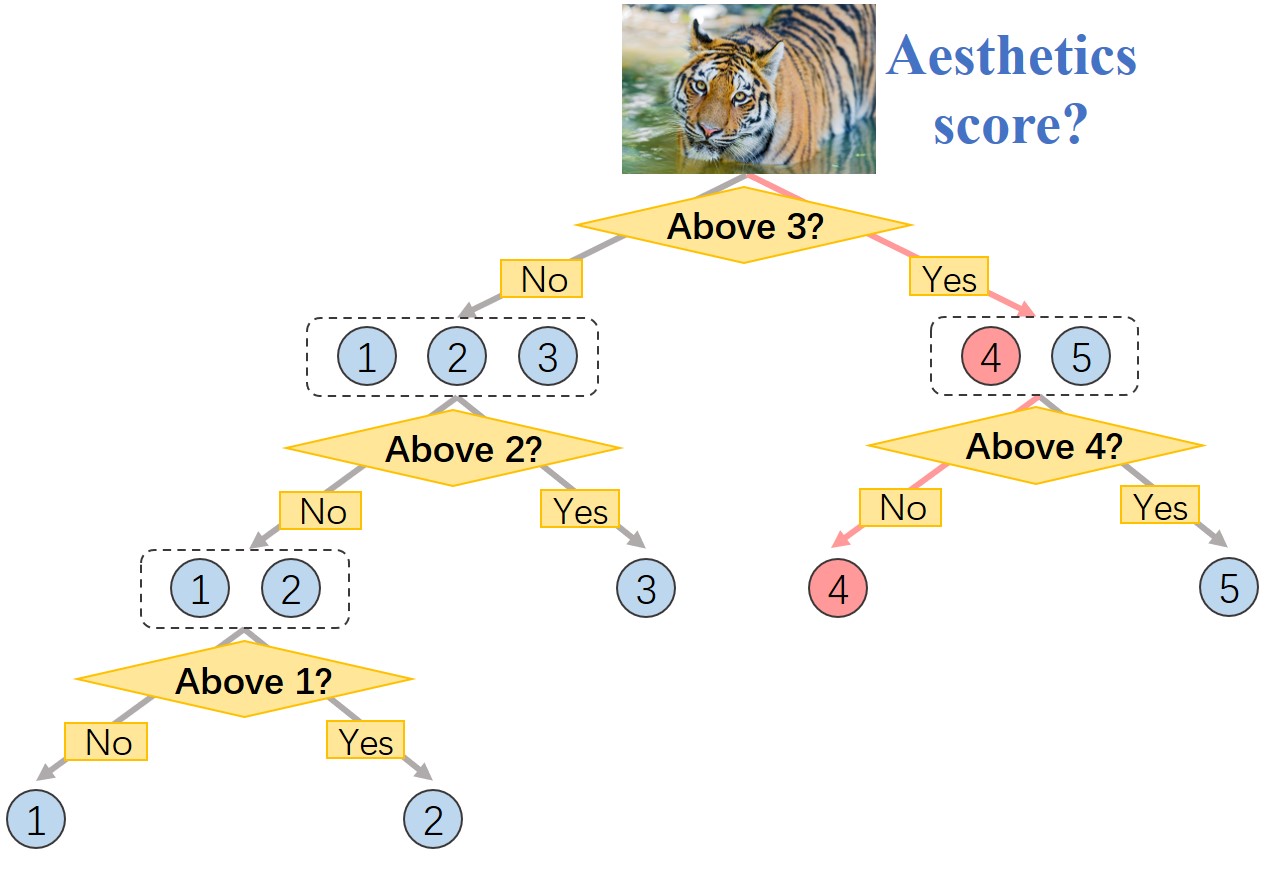}
\caption{Our motivation. The dichotomic search (binary search) aims to repeatedly divide the half portion of a sorted array to find the target item. It can be utilized in ordinal regression tasks since the ordinal candidate labels can be regarded as a finite sorted array. Thus, an ordinal regression task is decomposed into multiple recursive dichotomic classification sub-problems. For example, when scoring an aesthetic image (e.g., from 1 to 5, the ground truth is 4), we can first estimate whether the score is above or below average (i.e., \textcircled{3}). Next, if it is above average, then we can further determine the score to be \textcircled{4} or \textcircled{5}.}
\label{fig1}
\vskip -1 em
\end{figure}

In this paper, we argue and validate the importance of the ``adjacent categories distinction'' in ordinal regression tasks. To this end, we propose to distinguish the adjacent categories gradually in processing. Motivated by the dichotomic search (binary search)~\cite{williams1976modification}, which repeatedly divides the half portion of a sorted array to gradually find the target item, we decompose an ordinal regression problem into a series of dichotomic classification steps. \emph{In each step, we can only focus on dealing with a boundary of a pair of adjacent categories.}
An example is given in Fig.~\ref{fig1}. The aesthetics score of an image is gradually distinguished via recursive dichotomic classification.
In this way, an ordinal regression problem can be transformed into a sequence prediction problem that sequentially conducts dichotomic classification to finally obtain the ordinal category label.

Evolved from our motivation, we propose a simple sequence prediction framework for ordinal regression, called Ord2Seq. In our approach, ordinal regression is regarded as a sequence prediction task where the predicting goal is changed from a category label to a binary label sequence. That is, the prediction task is decomposed into a series of recursive binary classification steps to better distinguish adjacent categories in a process of progressive elaboration. Specifically, Ord2Seq performs two main steps. First, in pre-processing, we transform ordinal regression labels into label sequences by a tree-structured label mapping approach (we call the tree structure \emph{dichotomic tree} in this paper). Thus, for each input data, the prediction objective turns to a sequence of binary labels. Next, it predicts this label sequence {\it progressively} via an encoder-decoder structured Transformer architecture. The Transformer is allowed to integrate context information by delivering the earlier image features and prediction results for the next token prediction. Also, the Transformer adapts to any sequence prediction length, so that our model has strong scalability on different tasks with various numbers of categories. Further, to enable our model to focus on each binary decision when distinguishing the remaining categories, the Transformer decoder is designed with a masked decision strategy to suppress the loss interference of the eliminated categories. Comprehensive experiments validate the superiority of our proposed Ord2Seq that carefully distinguishes adjacent categories.

Our main contributions are summarized as follows:
\vspace{-5 pt}
\begin{itemize}[itemsep=-5pt]
\item For the first time, we propose to transform ordinal category labels as label sequences using a \emph{dichotomic tree}, so as to tackle an ordinal regression task as a sequence prediction task.
\item We propose a new sequence prediction framework for ordinal classification, called Ord2Seq, which effectively distinguishes adjacent categories with a process of progressive elaboration.
\item We design a novel decoder with a masked decision strategy to suppress the loss interference of the eliminated categories in order to focus on distinguishing the remaining categories.
\item Extensive experiments show the effectiveness of each component and that Ord2Seq performs better in distinguishing adjacent categories and achieves state-of-the-art performances on various image datasets.
\end{itemize}

\section{Related Work}
\label{sec:related}

\subsection{Ordinal Regression}

\begin{figure*}[t]
\centering
\includegraphics[width=0.8\textwidth]{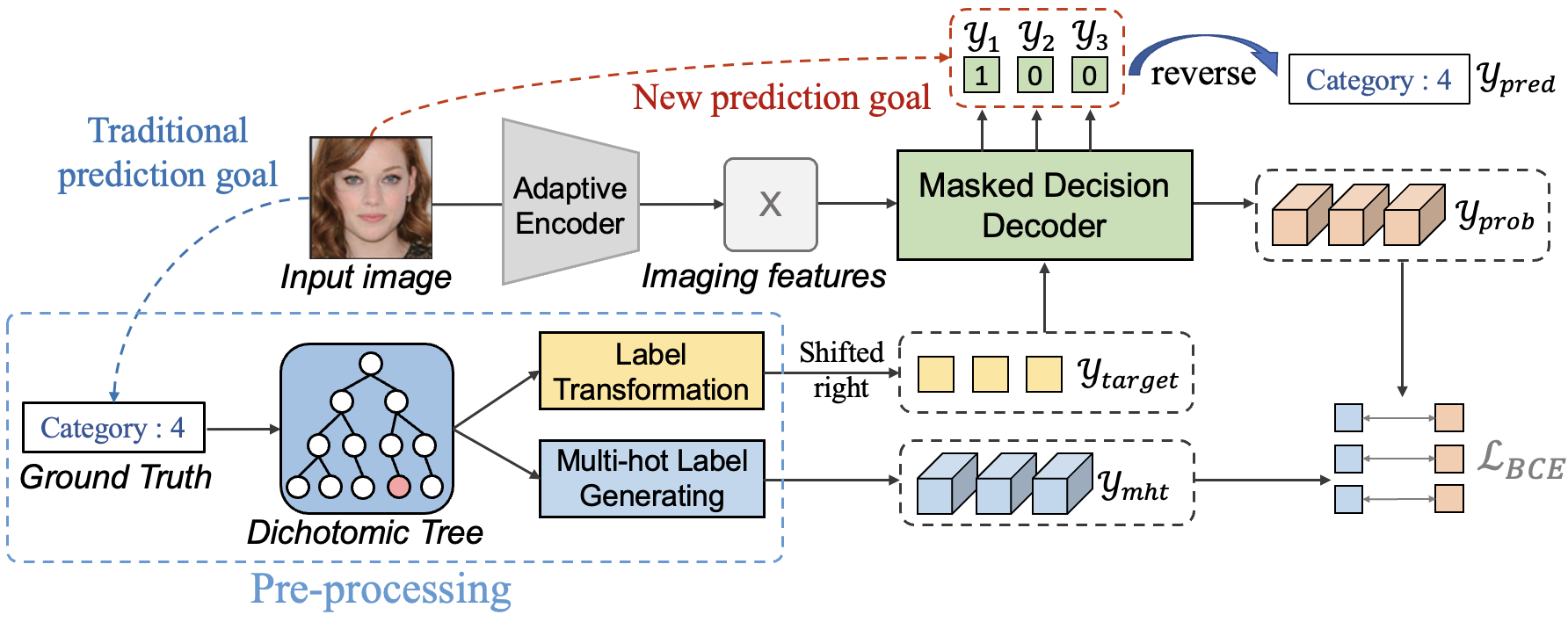}
\caption{An overview of our Ord2Seq approach. Given an input image (e.g., for aesthetic grading), Ord2Seq transforms ordinal category labels into a binary label sequence so that the prediction target becomes a label sequence rather than an independent category label.}
\label{fig2}
\vskip -0.2 in
\end{figure*}

The $K$-rank method~\cite{frank2001simple} is the most popular approach for ordinal regression, in which $K - 1$ classifiers are trained to rank ordinal categories. A study~\cite{li2006ordinal} combined mathematical analysis based on the $K$-rank method to better learn inter-class ordinal relationships. Some methods~\cite{niu2016ordinal,chen2017using} used trained convolutional neural networks as $K$-rank classifiers. Many recent studies~\cite{lim2019order, liu2019probabilistic, lee2020deep, li2021learning} proposed ordinal distribution constraints to exploit the ordinal nature of regression.
To add prior order knowledge to loss calculation, several methods~\cite{fu2018deep, diaz2019soft} created soft labels artificially by changing the distances between categories. A few advanced methods~\cite{liu2017deep, liu2018constrained, li2021learning, shin2022moving} sorted tuples that are formed by two~\cite{liu2018constrained} or three~\cite{liu2017deep, li2021learning, shin2022moving} instances with ordinal categories so the ranks of the test instances can be estimated from instances with known ranks. However, these methods only focused only on learning inter-class ordinal relationships and tend to be towards a misunderstanding that the latent features of adjacent categories should be as similar as possible. Consequently, these methods failed to highlight the boundaries between adjacent categories and perform not well in distinguishing adjacent categories, which hence hindered performance improvement. In this paper, we propose a dichotomy-based method to decompose ordinal regression into a series of recursive binary classification steps. With the candidate categories gradually refined, the model is able to focus on distinguishing adjacent categories. 



\subsection{Sequence Prediction}
Sequence prediction was first applied in the natural-language processing field (e.g., machine translation~\cite{sutskever2014sequence, bahdanau2014neural}). After Transformers~\cite{vaswani2017attention} were shown to have powerful capabilities in sequence prediction, many Transformer models were developed for sequence prediction~\cite{radford2019language, raffel2020exploring, brown2020language}, and were also gradually introduced to computer vision (CV)~\cite{dosovitskiy2020image}. But, in many CV tasks, Transformers were used only for feature extraction~\cite{liu2021swin, wang2021pyramid}. Inspired 
by the success of transforming different domains into sequence prediction~\cite{chen2021decision, ramesh2021zero}, a few studies treated CV tasks as sequence prediction~\cite{chen2021pix2seq, chen2022obj2seq}, and showed considerable effectiveness. In these methods, Transformers were used to not only extract features but also predict sequences that are related to the target CV tasks. Our work is also inspired by previous sequence prediction models based on the Transformer architecture. With the sequence prediction scheme, we achieve to bring the idea of dichotomic search into the ordinal regression task for the first time.

\section{Methodology}
\label{sec:methods}

\subsection{Overview}
Our proposed Ord2Seq model takes an image $I$ as input, and transforms the ground truth (ordinal category labels) into binary label sequences in order to regard ordinal regression as a sequence prediction task. Thus, the prediction goal becomes to output a sequence of binary labels, as shown in Fig.~\ref{fig2}. Ord2Seq consists of four main parts:
\vspace{-5 pt}
\begin{itemize}[itemsep=-5pt]
    \item Label Transformation and Multi-hot Label Generation: We construct a \emph{dichotomic tree} for pre-processing, which transforms ground truth (ordinal category labels) into a sequence of binary label, and then generates a sequence of multi-hot labels for loss calculation.
    \item Adaptive Encoder: We utilize an Adaptive Encoder to extract imaging features, which is compatible with both CNN and Transformer backbones.
    \item Masked Decision Decoder: Our Masked Decision Decoder can directly predict probability sequences and indirectly predict binary label sequences with a masked decision strategy (one token at a time).
    \item Loss Function: Our model is trained to minimize the sum of the binary cross-entropy (BCE) losses of matched pairs between predicted probability sequences and generated multi-hot label sequences.
\end{itemize}

\begin{figure*}[t]
\centering
\includegraphics[width=0.94\textwidth]{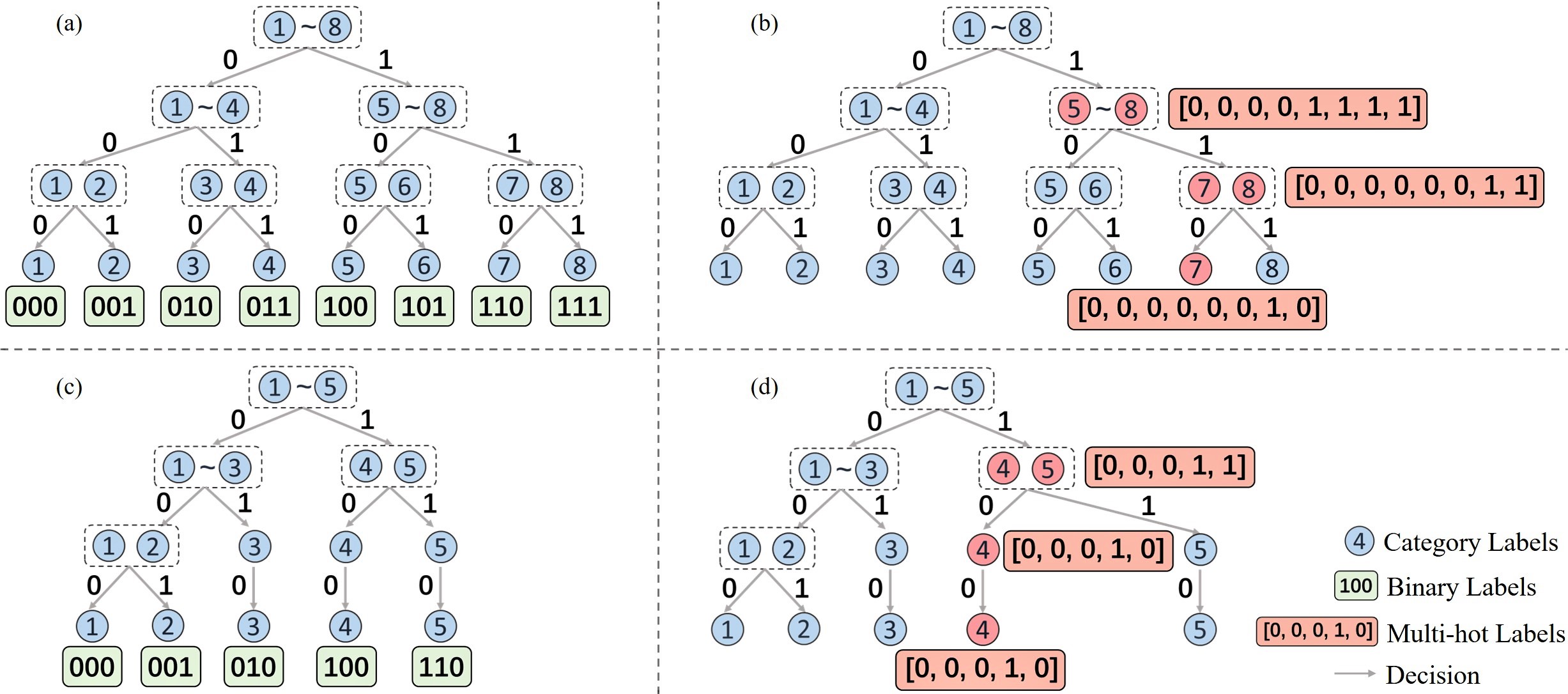}
\caption{Illustrating label transformation and multi-hot label generation via a \emph{dichotomic tree}. (a) Label transformation for $8=2^3$ categories. (b) Multi-hot label generation for 8 categories (taking category 7 as an example). (c) Label transformation for 5 categories. (d) Multi-hot label generation for 5 categories (taking category 4 as an example).}
\label{fig3}
\vskip -0.2 in
\end{figure*}
\subsection{Label Transformation and Multi-hot Label Generation}

\paragraph{Label Transformation via a \emph{Dichotomic Tree}.}
Based on the dichotomy algorithm, we design a \emph{dichotomic tree} to transform each ordinal category label into a sequence of binary label tokens for pre-processing. In this tree, the option paths to the left and right subtrees of each node are denoted by 0 and 1, respectively. If the number of categories is a power of 2, we construct a complete binary tree by dichotomy, as shown in Fig.~\ref{fig3}(a). However, when the number of categories is not a power of 2, we cannot ensure that the numbers of categories in the two subtrees of every node are the same. Therefore, we construct an incomplete dichotomic tree in which the left and right subtrees of every node do not differ by more than one node and the depths of the leaf nodes for each category are equal, as shown in Fig.~\ref{fig3}(c). After the tree construction, every category label $C$ is mapped to a corresponding binary label sequence $y_{b}$, showing an option path in the tree from the root node to the leaf node for the category $C$, by:
\begin{equation}
\centering
y_{b} = f(C) = [c_1, c_2,\ldots, c_d],
\end{equation}
\noindent where $c_i \in \{0, 1\}$ denotes the codes of the option path for the category $C$, and $d$ donates the height of the tree. 

Based on the constructed {\it dichotomic tree}, an ordinal category label is transformed into a binary label sequence, and our prediction target changes from a category label to a binary label sequence. Then, to predict the first label in sequence, following the \emph{shifted right} process in vanilla Transformer~\cite{vaswani2017attention}, we shift the binary label sequence $y_b$ right with a starting query token $s$:
\vspace{-6.5pt}
\noindent
\begin{equation}
\centering
y_{target}  = [s, c_1, c_2,\ldots, c_{d-1}].
\end{equation}

\vspace{-12pt}
\noindent
\paragraph{Multi-hot Label Generation.}
Different from the language models, we do not directly predict the binary label sequence since such binary labels may hinder the model prediction for two reasons. (1) The 0's and 1's at different positions in our binary sequence may have different meanings. Thus, the model cannot forecast them directly. (2) The scope and meaning of each binary classification are different, and the classifiers should differentiate. For (1), we use a \textit{Label Embedding} approach (presented in Section~\ref{sec:mdd}) to map different 0's and 1's into different embeddings. For (2), based on the built \emph{dichotomic tree}, we generate multi-hot label sequences which specify the scope and meaning of each classification. This process can be viewed as conducting continuous {\it range} predictions for the ground truth. Then the binary labels can be indirectly obtained from the range prediction results. Examples of \textit{Multi-hot Label Generation} are shown in Fig.~\ref{fig3}(b) and Fig.~\ref{fig3}(d). Each node of the tree corresponds to a multi-hot label, and every category label $C$ is mapped to a corresponding multi-hot label sequence, as: 
\begin{equation}
\centering
y_{mht} = g(C) = [o_1, o_2,\ldots, o_d],
\end{equation}
where $o_i = [o_{i, 1}, o_{i, 2},\ldots, o_{i, n}] $  denotes the multi-hot labels of the path for the category $C$ with $o_{i, j} \in \{0, 1\}$ and $n$ being the number of categories.
Thus, the multi-hot label at each node includes positive and negative classes, where the positive class is defined as the categories that the node includes and the negative class is for the other categories. With the supervision of multi-hot label sequences, the model can first predict a probability sequence and then output the binary label sequence based on the predicted probability sequence. 

\subsection{Masked Decision Decoder for Sequence Prediction}
\label{sec:mdd}

\begin{figure}[t]
\centering
  \includegraphics[width=0.238\textwidth]{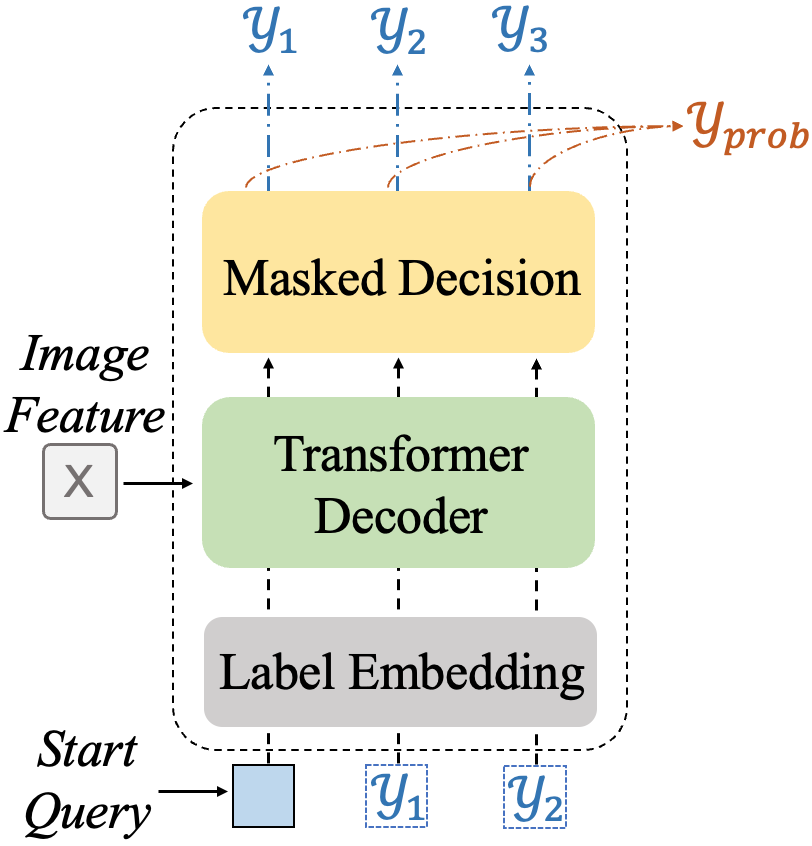}%
  \includegraphics[width=0.235\textwidth]{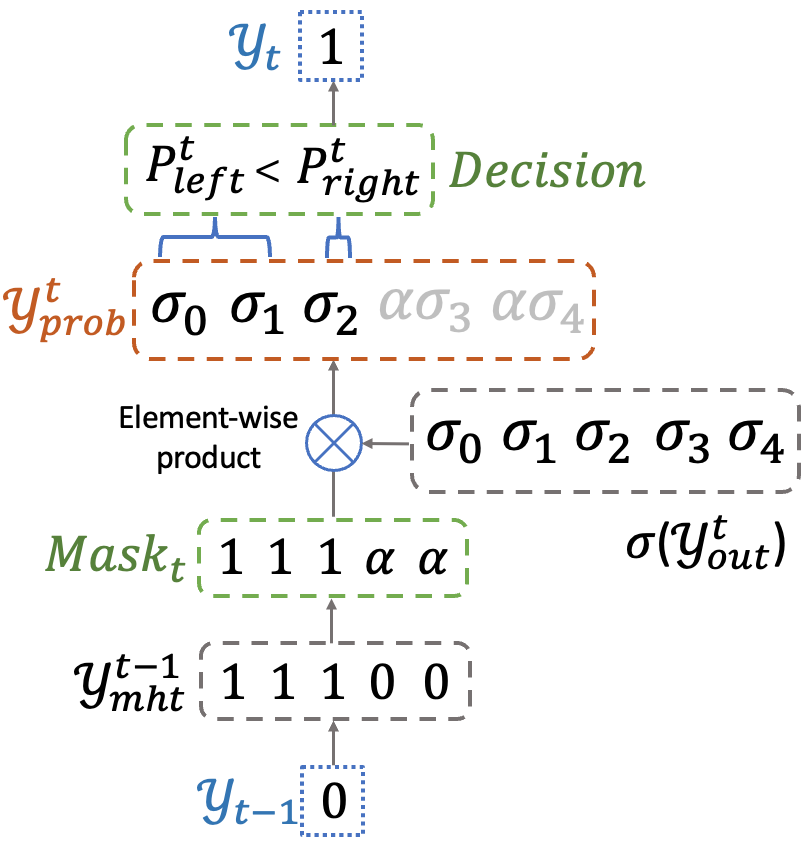}\\%
  (a) \quad \quad \quad \quad \quad \quad \quad \quad \quad (b)\\
\caption{(a) Illustrating the Masked Decision Decoder. (b) Illustrating how Masked Decision generates a probability label and a binary label at time step $t$. $Mask_{t}$, $y^{t}_{out}$, $y^{t-1}_{mht}$ and $y^{t}_{prob}$ correspond to Eq.(6) and Eq.(7). $\sigma$ denotes the sigmoid function. }
    \vskip -1 em
\label{fig4}
\end{figure}

The masked decision decoder takes imaging features $X$ obtained by the Adaptive Encoder and a target sequence $y_{target}$ as input, predicts a probability sequence $y_{prob}$, and outputs a binary label sequence $y$. Fig.~\ref{fig4}(a) overviews the masked decision decoder with its three main parts: Label Embedding, Transformer Decoder, and Masked Decision. 

\vspace{-8pt}
\noindent
\paragraph{Label Embedding.} To enable different 0's and 1's in each binary label sequence to represent different meanings, similar to the \textit{Position Encoding} in~\cite{dosovitskiy2020image}, we use a function $h$ to map the target binary label sequence $y_{target}$ to a new vector with different values, and then encode the vector to the embeddings $y_{embd}$ with the same size of Transformer tokens via an embedding layer $E$, which can be formulated as:
\textcolor{black}{
\vspace{-3pt}
\noindent
\begin{equation}
\centering
\begin{aligned}
    h(y_{target}) &= 2 \times i + y_{target} \\
    y_{embd} &= E(h(y_{target})). 
\end{aligned}
\end{equation}
}
\vspace{-18pt}
\noindent
\paragraph{Transformer Decoder.}
Our Transformer decoder $\mathcal{D}$ follows the vanilla architecture~\cite{vaswani2017attention} composed of Multi-headed Self-Attention (MSA), Layer Normalisation (LN), and Multi-headed Cross-Attention (MCA) layers with residual connections, aiming to predict the original logits sequence. For a time step $t$, the decoder $\mathcal{D}$ takes the $t^{th}$ embedding token $y_{embd}^{t}$ as the input query $y_{in}^{t}$ and then sent it to the MSA and MCA layers, and a linear layer $w_{t}$, in sequence, to finally produce the original logits $y_{out}^{t}$, where MSA takes the previous input $y_{in}^{1:t}$ to compute keys and values, and MCA takes imaging features $X$ for attention calculation. We formulate the process at time step $t$ as:
\vspace{-1.5pt}
\noindent
\begin{equation}
\centering
\begin{aligned}
y_{in}^{t}& =\left\{
	\begin{aligned}
	&y_{embd}^{t}   &\text{if training},\\
	&E(h({y_{t-1}})) &\text{if testing},\\
	\end{aligned}
	\right.\\
y_{hidden, 1}^{t}&=\text{LN}(\text{MSA}(y_{in}^{t}W_{Q};y_{in}^{1:t}W_{K};y_{in}^{1:t}W_{V})), \\
y_{hidden, 2}^{t} &= \text{LN}(\text{MCA}(y_{hidden, 1}^{t};X)), \\
y_{out}^{t} &= y_{hidden, 2}^{t}w_{t}^{T},\\
\end{aligned}
\end{equation}
where $W_{Q}$, $W_{K}$, and $W_{V}$ are weight matrices for computing queries, keys, and values.
The logits $y_{out}^t$ are then used to generate a probability prediction $y_{prob}^t$ and a binary label $y_t$ via the masked decision strategy (discussed below). Note that during testing, the decoder $\mathcal{D}$ takes the predicted binary label $y_{t-1}$ after {\it Label Embedding} as the input query $y_{in}^{t}$.
\vspace{-12pt}
\noindent
\paragraph{Masked Decision.}
The Masked Decision strategy is used to transform the original logits $y_{out}$ to a probability sequence $y_{prob}$ and predict a binary label sequence $y$ where the probability sequence $y_{prob}$ is for loss calculation and the binary label sequence $y$ is our prediction goal. By default, we perform the probability sequence prediction by $y_{prob} = {\rm sigmoid}(y_{out}$). But obviously, for each time step $t$, the prediction should be based on the previous results. Thus, we try to suppress the loss interference of the eliminated categories in the previous prediction (time step $t-1$) with a mask. As shown in Fig.~\ref{fig4}(b), for a time step $t$, the mask is defined as:
\vspace{-6pt}
\noindent
\begin{equation}
\centering
\text{Mask}_{t,i} =\left\{
	\begin{aligned}
	&1   \quad &y_{mht}^{t-1,i}=1,\\
	&\alpha \quad &y_{mht}^{t-1,i}=0,\\
	\end{aligned}
	\right.
\end{equation}
where $\alpha$ is a hyper-parameter (we set $\alpha = 0.3$). Then the probability prediction at time step $t$ becomes:
\begin{equation}
\centering
y_{prob}^{t} = \text{Mask}_t \odot \text{sigmoid}(y_{out}^{t}),
\end{equation}
where $\odot$ is the element-wise product. Since $\alpha < 1$, the mask can be used to reduce the probability value of the $i^{th}$ category that satisfies $y_{mht}^{t,i}=0$ (because all such categories have been eliminated in previous steps). Hence, the loss interference of these eliminated categories is restrained when calculating the loss between the predicted probability sequence $y_{prob}$ and the multi-hot sequence $y_{mht}$, forcing the model to focus on distinguishing the remaining categories.

After the masking process, we apply a decision strategy to predict the binary label based on the unmasked categories in $y_{prob}^{t}$ (see Fig.~\ref{fig4}(b)). Suppose the categories of the left subtree are in $[l, m]$ and the categories of the right subtree are in $[m+1, r]$. We compute the average of all the probability values in each subtree, and compare them. According to the comparison result, we obtain the binary label $y_t$ for time step $t$. This process can be formulated as:
\begin{equation}
\centering
\begin{aligned}
P_{left}^{t} &=\frac{1}{m-l+1}\sum_{i=l}^{m}y_{prob}^{t, i}, \\
P_{right}^{t} &=\frac{1}{r-m}\sum_{i=m+1}^{r}y_{prob}^{t, i}, \\
{y}_{t}& =\left\{
	\begin{aligned}
	&0   &P_{left}^{t} &\geq P_{right}^{t},\\
	&1   &P_{left}^{t} &< P_{right}^{t},\\
	\end{aligned}
	\right.\\
\end{aligned}
\end{equation}
where $P_{left}^{t}$ and $P_{right}^{t}$ denote the average of the probability values of the categories in the left and right subtrees, respectively. The obtained binary label will be used for the next label prediction. As more binary labels are predicted, the remaining candidate categories are gradually dwindling and the adjacent categories are finally distinguished with higher confidence. After all the steps, we can inverse-map the resulted binary label sequence to the true category:
\vspace{-6pt}
\begin{equation}
\centering
y_{pred} = f^{-1}(y).
\end{equation}

It can be seen that our masked decision decoder effectively joins the sequence prediction and decision-making process by first predicting a probability sequence via the Transformer decoder and then predicting a binary label sequence via the masked decision strategy. 

  
 



\subsection{Other Details}

\paragraph{Adaptive Encoder.} Our plug-and-play method adapts to any encoder-decoder architecture. Most existing vision Transformers are suitable as our encoder. In this work, we choose PVTv2~\cite{wang2022pvt} as the encoder. Further, to adapt to popular CNN encoders such as VGG~\cite{simonyan2014very}, we follow~\cite{carion2020end} by flattening the feature map after stage 5 (DC5); then the feature map is transformed to 512 channels and is passed to a Transformer encoder to obtain the imaging features $X$.
\vspace{-10pt}
\noindent
\paragraph{Loss Functions.} Unlike the commonly used Cross Entropy (CE) loss in most ordinal classification methods, we choose Binary Cross Entropy (BCE) loss since our multi-hot labels have multiple positive classes which can be regarded as a multi-label classification problem for which CE loss is not suitable while BCE loss is. We first calculate the BCE loss between $y_{prob}^{t}$ and $y_{mht}^{t}$ at each time step $t$, and then sum them up as:
\begin{equation}
\begin{aligned}
    L = \sum_{t=1}^{d}BCE(y_{prob}^{t},y_{mht}^{t})=-\frac{1}{n}\sum_{t=1}^{d}\sum_{i=1}^{n}\\
    (y_{mht}^{t, i}\log(y_{prob}^{t, i})
      +(1-y_{mht}^{t, i})\log(1-y_{prob}^{t, i})).
\end{aligned}
\end{equation}

\section{Experiments}
To validate the effectiveness of our Ord2Seq approach, we conduct extensive experiments on the datasets of four different scenarios: Image Aesthetics, Age Estimation, Historical Image Dating, and Diabetic Retinopathy Grading.

\subsection{Experimental Setup}
Our experiments use a computer with an Intel i7 processor and an NVIDIA GTX 2080Ti GPU. To compare with existing methods that use VGG-16 as the backbone, we train Ord2Seq with two Adaptive Encoders, VGG-16~\cite{simonyan2014very} and PVTv2-b1~\cite{wang2022pvt}, with similar settings and pre-trained on ImageNet~\cite{russakovsky2015imagenet}. The mini-batch size is 32. We use random horizontal flipping and random cropping to the crop size of 224 $\times$ 224 for data augmentation. For optimization, the Adam optimizer~\cite{kingma2014adam} is utilized with a learning rate of $10^{-4}$. For fair comparisons, all the known methods are implemented using the authors’ code or re-implemented based on the original papers. More details about the datasets and experimental settings are in the supplemental document.

\subsection{Age Estimation}
\paragraph{Dataset:} The Adience dataset~\cite{levi2015age} is used for age group estimation that contains about 26,580 face images from Flickr of 2,284 subjects. Ages are annotated in 8 groups: 0-2, 4-6, 8-13, 15-20, 25-32, 38-43, 48-53, and over 60 years old. All the images are divided into 5 subject-exclusive folds for cross-validation as in~\cite{liu2018constrained, diaz2019soft, li2021learning, shin2022moving}.

\begin{figure}[t]
  \centering
  \includegraphics[width=0.445\textwidth]{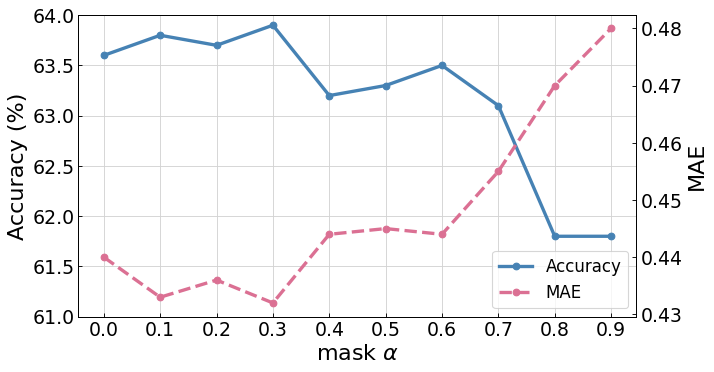}

   \caption{Ord2Seq (PVT) performances with different values of the mask $\alpha$ {on the Adience dataset}. It achieves the best performance when $\alpha = 0.3$.}
   \label{fig:5}
\end{figure}

\vspace{-12pt}
\noindent
\paragraph{Mask Parameter Analysis:} We first explore the effects of different values of the mask $\alpha$ on the Adience dataset. Fig.~\ref{fig:5} shows the results for $\alpha$ varying from 0 to 0.9 with an interval of 0.1. One can see that Ord2Seq attains the best performance when $\alpha = 0.3$, which shows that it is helpful to suppress the loss interference of the eliminated categories, thus letting the classifiers focus on the discrimination of the remaining categories. 
In addition, we find that the model performance decreases slightly when $\alpha < 0.3$, which suggests that the model prefers a soft suppression since the sharp loss suppression may destabilize the model. In the following experiments, we set the mask value $\alpha=0.3$.



\vspace{-12pt}
\noindent
\paragraph{Ablation Study:}  We conduct an ablation study to evaluate the effectiveness of the sequence prediction scheme by keeping the Transformer decoder structure but removing the sequence prediction scheme to only predict a true label once (VGG+Trans) and evaluate the masked decision strategy by removing the masking process (Ord2Seq (VGG)\dag). Table~\ref{tab:11} shows the results. Compared to the `VGG only' baseline, the `VGG + Trans' model improves performance slightly (by +0.4\% on accuracy and -0.4 on MAE) while the superiority of our proposed sequence prediction model Ord2Seq is significant (achieving accuracy increases by 4.2\%). This result indicates that the core reason of our performance gain is the proposed sequence prediction scheme, rather than the larger Transformer network. Besides, our masked decision strategy further considerably improves the performance (accuracy increases by 6.1\%), demonstrating the effectiveness of our mask design to suppress the loss interference from the eliminated categories that promote the distinction of the remaining categories.


\begin{table}[t]
\centering
\begin{tabular}{lll}
\toprule
Method & Accuracy (\%) & MAE  \\ \midrule
VGG only      & \quad 57.4          & \ 0.55\\ 
VGG + Trans   & \quad 57.8\quad(\textcolor{dg}{+0.4})        & \ 0.51 \quad (\textcolor{dg}{-0.04})\\ 
Ord2Seq (VGG)\dag & \quad 61.6\quad(\textcolor{dg}{+4.2})      & \ 0.49 \quad (\textcolor{dg}{-0.06})\\ 
Ord2Seq (VGG) & \quad {\bf 63.5}\quad(\textcolor{dg}{+6.1})     & \ {\bf 0.44} \quad (\textcolor{dg}{-0.11})\\ \bottomrule
\end{tabular}%
\caption{Ablation experiments on the Adience dataset. For Accuracy, higher is better; for MAE, lower is better. \dag denotes the Ord2Seq model without the masked decision strategy.}
\label{tab:11}
\end{table}

\begin{table}
  \centering
 
  \begin{tabular}{@{\quad}lccc@{\quad}}
    \toprule
    Method & Accuracy (\%) & MAE & Params. \\
    \midrule
    Lean DNN~\cite{levi2015age} & 50.7 & -- & -- \\
    Niu et al.~\cite{niu2016ordinal} & 56.7 & 0.55 & -- \\
    CNNPOR~\cite{liu2018constrained} &  57.4 & 0.55 & --  \\
    GP-DNNOR\cite{liu2019probabilistic} & 57.4 & 0.54 & -- \\
    SORD~\cite{diaz2019soft} & 59.6 & 0.49 & 138.4M \\
    POE~\cite{li2021learning} & 60.5 & 0.47 & 151.1M\\
    MWR~\cite{shin2022moving} & 62.6 & 0.45 & 597.0M \\
    \midrule
    Ours (VGG) & {\underline{63.5}} & {\underline{0.44}} & 182.8M  \\ 

    Ours (PVT) & {\bf 63.9} & {\bf 0.43} & 187.5M \\

    \bottomrule
  \end{tabular}

  \caption{Accuracy and MAE comparison on the Adience dataset.}
  \label{tab:2}
  \vskip -1 em
\end{table}

\begin{table*}
  \centering

  \begin{tabular*}{16.8cm}{lcccccccccc}
    \toprule
     \multirow{2}*{Method} &
    \multicolumn{5}{c}{Accuracy (\%) -- higher is better} 
    &  \multicolumn{5}{c}{MAE -- lower is better}
    \\
    \cmidrule(r){2-6} \cmidrule(r){7-11}  & Nature & Animal & Urban & People & Overall 
    & Nature & Animal & Urban & People & Overall
    \\
    \midrule
    CNNPOR~\cite{liu2018constrained} & 71.86 & 69.32 & 69.09 & 69.94 & 70.05 & 0.294 & 0.322 & 0.325 & 0.321 & 0.316 \\
    SORD~\cite{diaz2019soft} & 73.59 & 70.29 & {\underline{73.25}} & {\underline{70.59}} & 72.03 & 0.271 & 0.308 & 0.276 & {\underline{0.309}} & 0.290 \\
    POE~\cite{li2021learning}  & 73.62 & 71.14 & 72.78 & {\bf 72.22} & 72.44 & 0.273 & 0.299 & 0.281 & {\bf 0.293} & 0.287\\
   
    \midrule
    Ours (VGG) & {\bf 78.22} & {\underline{73.77}} & {\bf 73.57} & 68.69 & {\underline{74.02}} & {\bf 0.221}	& {\underline{0.271}} & {\bf 0.270} & 0.326 & {\underline{0.267}} \\ 

    Ours (PVT) & {\underline{78.09}}	& {\bf 75.74} & 72.83 & 69.24 & {\bf 74.43} & {\underline{0.225}} & {\bf 0.257} & {\underline{0.275}} & 0.319 & {\bf 0.264} \\

    \bottomrule
  \end{tabular*}

  \caption{Results on the Image Aesthetics dataset. Accuracy and MAE are reported for each of the four image classes.}
  \label{tab:3}
\end{table*}

\begin{table}
  \centering
 
  \begin{tabular}{@{\quad}lcc@{\quad}}
    \toprule
    Method & Accuracy (\%) & MAE \\
    \midrule
    Palermo et al.~\cite{palermo2012dating} & 44.9 $\pm$ 3.7 & 0.93 $\pm$ 0.08\\ 
    CNNPOR~\cite{liu2018constrained}  & 50.1 $\pm$ 2.7 & 0.82 $\pm$ 0.05 \\
    GP-DNNOR\cite{liu2019probabilistic}  & 46.6 $\pm$ 3.0 & 0.76 $\pm$ 0.05\\
    SORD~\cite{diaz2019soft}   & 53.4 $\pm$ 3.7 & 0.70 $\pm$ 0.05\\
    POE~\cite{li2021learning}  & 54.7 $\pm$ 3.2 & 0.66 $\pm$ 0.05\\
    MWR~\cite{shin2022moving}  & 57.8 $\pm$ 4.1 & 0.58 $\pm$ 0.05\\
    \midrule
    Ours (VGG)  & {\underline{59.5 $\pm$ 1.7}} & {\underline{0.53 $\pm$ 0.03}}  \\ 

    Ours (PVT)  & {\bf 60.9 $\pm$ 1.6} & {\bf 0.52 $\pm$ 0.01}  \\

    \bottomrule
  \end{tabular}

  \caption{Accuracy and MAE comparison on the HCI dataset.}
  \label{tab:5}
   \vskip -1 em
\end{table}

\vspace{-14pt}
\noindent
\paragraph{Comparison with Known Methods:} We show the comparison results on the Adience dataset in Table~\ref{tab:2}. We find that our Ord2Seq with the VGG encoder achieves better results than the existing methods that use the same VGG architecture. Compared with POE~\cite{li2021learning} and SORD~\cite{diaz2019soft}, in despite slightly larger size, Ord2Seq achieves significant performance gains. Compared with the previous SOTA model MWR~\cite{shin2022moving}, our proposed Ord2Seq achieves a superior performance while the model sizes are much smaller than MWR. These comparison results demonstrate the superiority of our proposed approach by focusing on distinguishing adjacent categories. Further, Ord2Seq with the PVT encoder achieves an accuracy of 63.9\% and MAE of 0.43, defeating all current state-of-the-art results, which shows that using a Transformer encoder as the backbone can exert the potential of our approach due to its powerful representation capabilities.

\subsection{Image Aesthetics}
\paragraph{Dataset:} The Aesthetics dataset~\cite{dosovitskiy2020image} contains 15,687 Flickr image URLs, 13,706 of which are available. The dataset is used to grade image aesthetics. There are four image classes: animals, urban, people, and nature. Each image was graded by at least 5 different graders in 5 ranking categories to evaluate the photographic aesthetic quality: unacceptable, flawed, ordinary, professional, and exceptional. The ground truth is defined as the median rank among all the gradings. Following~\cite{liu2018constrained,diaz2019soft, li2021learning}, we apply 5-fold cross-validation. The images are randomly divided by 75\%, 5\%, and 20\% for training, validation, and testing, respectively.

\vspace{-12pt}
\noindent
\paragraph{Results:} Table~\ref{tab:3} shows the results on the Image Aesthetics dataset. We observe that our approach significantly outperforms the existing methods in various metrics. For example, our model with the PVT encoder achieves an Accuracy of 78.09\% for the Nature class, an overall Accuracy of 74.43\%, and an overall MAE of 0.264, outperforming the POE method~\cite{li2021learning} by 4.47\% for the Nature class, 1.99\% for the overall Accuracy, and 0.023 for the overall MAE. Except the mediocre performances of our model for the People class that may be due to that people's aesthetics is more subjective with various factors (e.g., gender, age, expression, etc), Ord2Seq achieves state-of-the-art results for the other classes, which validate the effectiveness of our approach.



\subsection{Historical Image Dating}
\paragraph{Dataset:} The historical color image (HCI) dataset is for estimating the decades of historical color photos. There are five decades from 1930s to 1970s annotated as 1 to 5. Each decade has 265 images. Following~\cite{liu2018constrained, liu2019probabilistic, li2021learning, shin2022moving}, we randomly split the 265 images of each decade into three subsets: 210 for training, 5 for validation, and 50 for testing. Then 10-fold cross-validation is performed, and the mean values of the results are recorded.

\begin{figure}[t]
\centering
\includegraphics[width=0.48\textwidth]{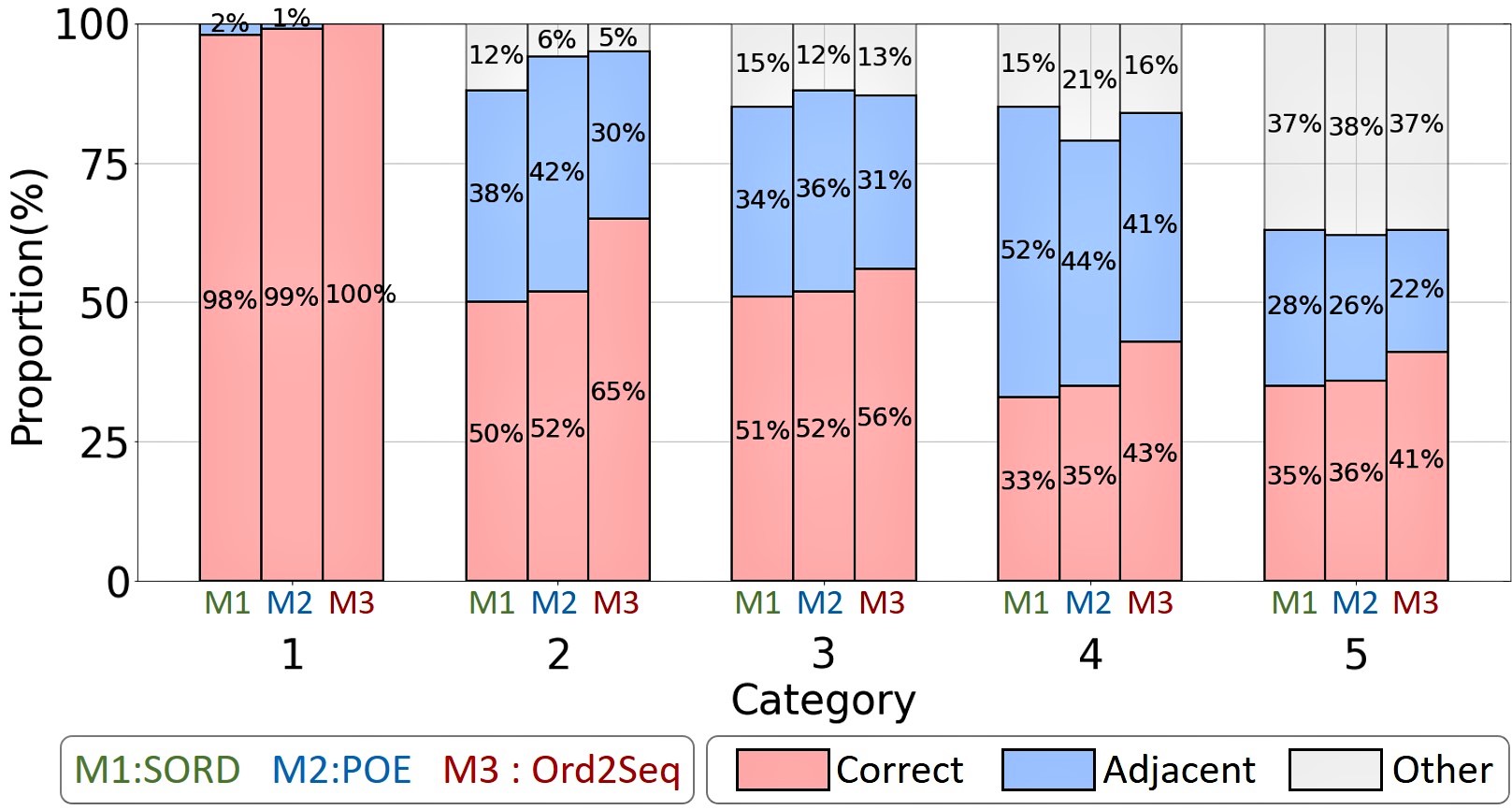}
\caption{Performances of SORD~\cite{diaz2019soft}, POE~\cite{li2021learning}, and Ord2Seq (PVT) for each category on the HCI dataset, showing the proportions of samples that belong to one category and are predicted to the correct, adjacent, and other categories. For example, for all samples whose ground truth are category 3, M3 (Ord2Seq) predicted 56\% samples correctly, 31\% samples to the adjacent categories (2 or 4), and 13\% samples to the other categories (1 or 5).} 
\label{fig7}
\vskip -1 em
\end{figure}

\vspace{-12pt}
\noindent
\paragraph{Results:} Table~\ref{tab:5} compares the results on the HCI dataset. As can be seen, our Ord2Seq with the VGG encoder outperforms known methods that use the same VGG architecture. Further, Ord2Seq with the PVT encoder achieves state-of-the-art results, providing improvements of 3.1\% in Accuracy and 0.06 in MAE, which indicate the superiority of our approach. In addition, to provide more details of the model performances in distinguishing adjacent categories, for all samples whose ground truths are of the same category, we calculate the proportions of these samples that are predicted to the correct, adjacent, and other categories. As visualized in Fig.~\ref{fig7}, we find that, for most categories, although the sums of the correct and adjacent proportions attained by different methods are close, our proposed Ord2Seq achieves higher proportions of the correct predictions and lower proportions of the adjacent predictions. That is, Ord2Seq is able to successfully predict part of samples that tended to be predicted into adjacent categories by previous methods.  This result shows the effectiveness of Ord2Seq in distinguishing adjacent categories, which is also the main performance improvement of our method comes from. 
 
\begin{figure*}[t]
\centering
\includegraphics[width=0.8\textwidth]{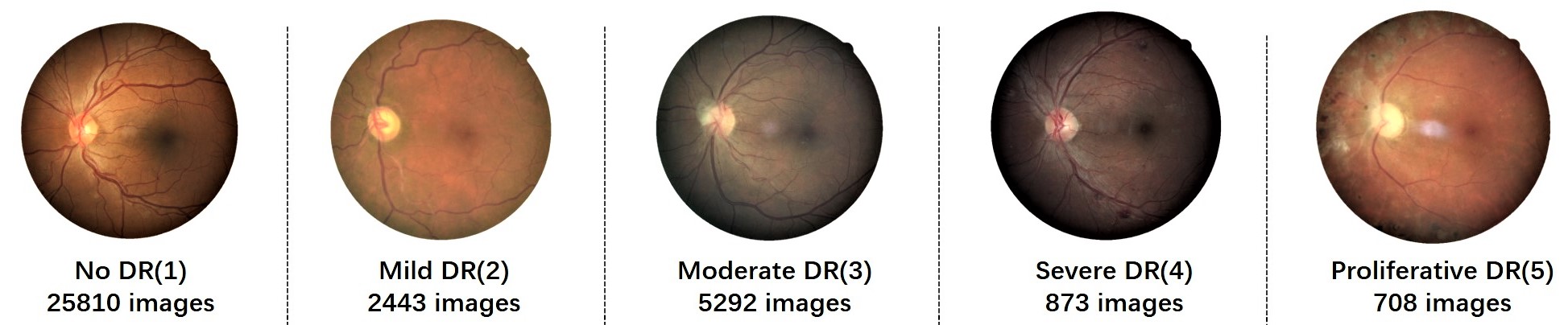}
\caption{Some sample fundus images with different diabetic retinopathy levels in the DR dataset.}
\label{fig8}
\vskip -1 em
\end{figure*}

\subsection{Diabetic Retinopathy Grading}
\paragraph{Dataset:} The Diabetic Retinopathy (DR) dataset contains 35,126 high-resolution fundus images available at \href{https://www.kaggle.com/c/diabetic-retinopathy-detection}{https://www.kaggle.com/c/diabetic-retinopathy-detection}. In this dataset, images were annotated in five levels of diabetic retinopathy from 1 to 5, representing no DR (25,810 images), mild DR (2,443 images), moderate DR (5,292 images), severe DR (873 images), and proliferative DR (708 images), respectively. Some sample images are shown in Fig.~\ref{fig8}. Following the setting used in~\cite{beckham2017unimodal, liu2018ordinal}, we apply the subject-independent 10-fold cross-validation, and report the mean values of the results.

\begin{table}
  \centering
 
  \begin{tabular}{@{\quad}lcc@{\quad}}
    \toprule
    Method & Accuracy (\%) & MAE \\
    \midrule
    Poisson~\cite{beckham2017unimodal} & 77.1 $\pm$ 0.6 & 0.38 $\pm$ 0.25 \\ 
    MT~\cite{ratner2018learning} & 82.8 $\pm$ 0.6 & 0.36 $\pm$ 0.22 \\

    SORD~\cite{diaz2019soft} & 78.2 $\pm$ 0.6 & 0.73 $\pm$ 0.17  \\
    POE~\cite{li2021learning} & 80.5 $\pm$ 0.6 & 0.30 $\pm$ 0.21  \\
 
    \midrule
    Ours (VGG) & {\underline{84.0 $\pm$ 0.6}} & {\underline{0.25 $\pm$ 0.07}}   \\ 

    Ours (PVT) & {\bf 84.2 $\pm$ 0.5} & {\bf 0.25 $\pm$ 0.07}   \\

    \bottomrule
  \end{tabular}

  \caption{Accuracy and MAE comparison on the DR dataset.}
  \label{tab:4}
\vskip -1 em
\end{table}

\begin{figure}[t]
\centering
\includegraphics[width=0.48\textwidth]{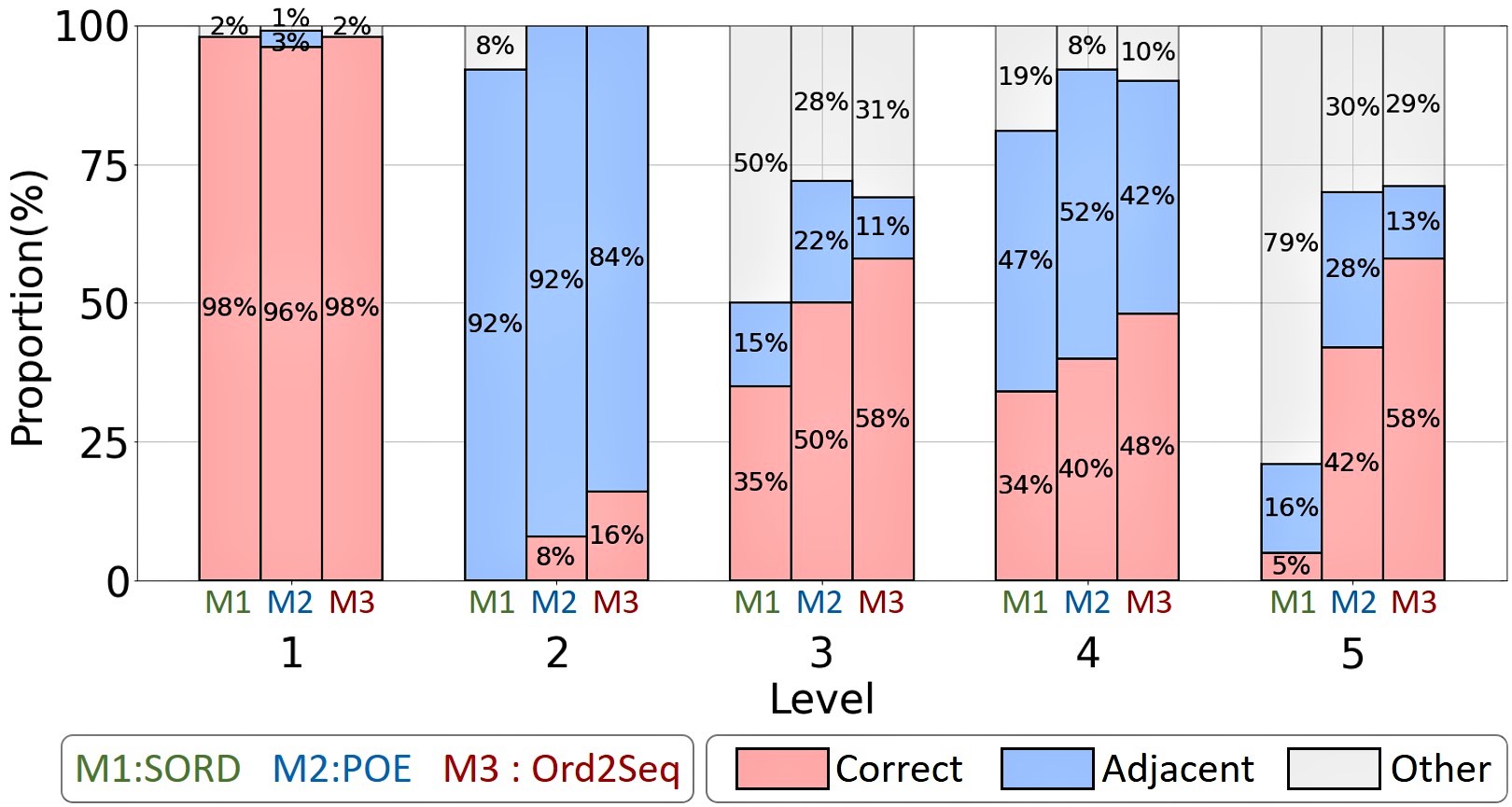}
\caption{Performances of SORD, POE, and Ord2Seq (PVT) for each category on the DR dataset, showing the proportions of samples that truly belong to one level and are predicted to the correct, adjacent, and other levels. Although overall performance is still limited on unbalanced categories, Ord2Seq significantly improves the Accuracy performance in distinguishing adjacent categories.}

\label{fig9}
\vskip -1 em
\end{figure}

\vspace{-14pt}
\noindent
\paragraph{Results:} Table~\ref{tab:4} shows the results on the DR dataset. Note that the DR dataset is unbalanced since the sample number decreases sharply as the severity DR level increases. We observe that the known order learning methods yield poor performances which may be due to the unbalanced data. Especially, SORD~\cite{diaz2019soft}, which is a modality-specific method by utilizing modified soft labels, can suffer serious errors in MAE. In comparison, our proposed Ord2Seq still maintains competitive performances, achieving an Accuracy of 84.2\% and an MAE of 0.25, which greatly outperforms the baselines and the other order learning methods, showing that our approach has better robustness on unbalanced data. We believe that this is due to the better positive-negative distinction. That is, unlike one positive class against other negative classes in previous work, it turns to (e.g.) classifier the first two categories against last three categories in the first step of Ord2Seq for the DR dataset (5 categories in total). In this way, the classification in a step is more category-balanced and helps to better exert unbalanced data. Moreover, to validate the superiority of Ord2Seq in distinguishing adjacent categories, we visualize the model performances based on the proportions of samples that truly belong to one category and are predicted to the correct, adjacent, and other categories on the DR dataset, in Fig.~\ref{fig9}. It is obvious that for levels 2--5 (with limited numbers of samples), although all the compared methods yield sub-optimal performances, our Ord2Seq significantly improves the correct prediction proportions and reduces the adjacent prediction proportions. This result also validates that our approach has better generalization on unbalanced categories, and can effectively distinguish adjacent categories to achieve higher overall performance.



\vspace{-6pt}
\noindent

\vspace{-10pt}
\section{Conclusions}
In this paper, we proposed a new sequence prediction framework for ordinal regression, Ord2Seq, which transforms ordinal labels as binary label sequences and uses a dichotomy-based sequence prediction procedure to distinguish adjacent categories based on a progressive elaboration scheme. Extensive experiments showed that Ord2Seq achieves state-of-the-art performances in various applied scenarios, and verified that Ord2Seq can effectively distinguish adjacent categories for performance improvement. 

The insight of our approach, i.e. dichotomy-based sequence prediction, is instructive for other general classification tasks. By dividing similar categories into a subtree and gradually refining the classification through a sequence prediction process, our model may be able to effectively distinguish similar objects with fine-grained differences (e.g., donkeys and horses).
\vspace{-12pt}
\paragraph{Acknowledgements.} This research was partially supported by National Key R\&D Program of China under grant No. 2018AAA0102102, National Natural Science Foundation of China under grants No. 62176231, No. 82202984, No. 62106218, No. 92259202, No. 62132017 and U22B2034, Zhejiang Key R\&D Program of China under grant No. 2023C03053.

{\small
\bibliographystyle{ieee_fullname}
\bibliography{egbib}

\begin{thebibliography}{10}\itemsep=-1pt

\bibitem{bahdanau2014neural}
Dzmitry Bahdanau, Kyunghyun Cho, and Yoshua Bengio.
\newblock Neural machine translation by jointly learning to align and
  translate.
\newblock {\em arXiv preprint arXiv:1409.0473}, 2014.

\bibitem{beckham2017unimodal}
Christopher Beckham and Christopher Pal.
\newblock Unimodal probability distributions for deep ordinal classification.
\newblock In {\em International Conference on Machine Learning}, pages
  411--419. PMLR, 2017.

\bibitem{brown2020language}
Tom Brown, Benjamin Mann, Nick Ryder, Melanie Subbiah, Jared~D Kaplan, Prafulla
  Dhariwal, Arvind Neelakantan, Pranav Shyam, Girish Sastry, Amanda Askell,
  et~al.
\newblock Language models are few-shot learners.
\newblock {\em Advances in Neural Information Processing Systems},
  33:1877--1901, 2020.

\bibitem{carion2020end}
Nicolas Carion, Francisco Massa, Gabriel Synnaeve, Nicolas Usunier, Alexander
  Kirillov, and Sergey Zagoruyko.
\newblock End-to-end object detection with transformers.
\newblock In {\em European Conference on Computer Vision}, pages 213--229.
  Springer, 2020.

\bibitem{chen2021decision}
Lili Chen, Kevin Lu, Aravind Rajeswaran, Kimin Lee, Aditya Grover, Misha
  Laskin, Pieter Abbeel, Aravind Srinivas, and Igor Mordatch.
\newblock Decision {Transformer}: Reinforcement learning via sequence modeling.
\newblock {\em Advances in Neural Information Processing Systems},
  34:15084--15097, 2021.

\bibitem{chen2017using}
Shixing Chen, Caojin Zhang, Ming Dong, Jialiang Le, and Mike Rao.
\newblock Using ranking-{CNN} for age estimation.
\newblock In {\em Proceedings of the IEEE Conference on Computer Vision and
  Pattern Recognition}, pages 5183--5192, 2017.

\bibitem{chen2021pix2seq}
Ting Chen, Saurabh Saxena, Lala Li, David~J Fleet, and Geoffrey Hinton.
\newblock Pix2seq: A language modeling framework for object detection.
\newblock {\em arXiv preprint arXiv:2109.10852}, 2021.

\bibitem{chen2022obj2seq}
Zhiyang Chen, Yousong Zhu, Zhaowen Li, Fan Yang, Wei Li, Haixin Wang, Chaoyang
  Zhao, Liwei Wu, Rui Zhao, Jinqiao Wang, et~al.
\newblock Obj2seq: Formatting objects as sequences with class prompt for visual
  tasks.
\newblock {\em arXiv preprint arXiv:2209.13948}, 2022.

\bibitem{chylack1993lens}
Leo~T Chylack, John~K Wolfe, David~M Singer, M~Cristina Leske, Mark~A
  Bullimore, Ian~L Bailey, Judith Friend, Daniel McCarthy, and Suh-Yuh Wu.
\newblock The lens opacities classification system {III}.
\newblock {\em Archives of Ophthalmology}, 111(6):831--836, 1993.

\bibitem{diaz2019soft}
Raul Diaz and Amit Marathe.
\newblock Soft labels for ordinal regression.
\newblock In {\em Proceedings of the IEEE/CVF Conference on Computer Vision and
  Pattern Recognition}, pages 4738--4747, 2019.

\bibitem{dosovitskiy2020image}
Alexey Dosovitskiy, Lucas Beyer, Alexander Kolesnikov, Dirk Weissenborn,
  Xiaohua Zhai, Thomas Unterthiner, Mostafa Dehghani, Matthias Minderer, Georg
  Heigold, Sylvain Gelly, et~al.
\newblock An image is worth 16x16 words: Transformers for image recognition at
  scale.
\newblock {\em arXiv preprint arXiv:2010.11929}, 2020.

\bibitem{frank2001simple}
Eibe Frank and Mark Hall.
\newblock A simple approach to ordinal classification.
\newblock In {\em European Conference on Machine Learning}, pages 145--156.
  Springer, 2001.

\bibitem{fu2018deep}
Huan Fu, Mingming Gong, Chaohui Wang, Kayhan Batmanghelich, and Dacheng Tao.
\newblock Deep ordinal regression network for monocular depth estimation.
\newblock In {\em Proceedings of the IEEE Conference on Computer Vision and
  Pattern Recognition}, pages 2002--2011, 2018.

\bibitem{kingma2014adam}
Diederik~P Kingma and Jimmy Ba.
\newblock Adam: A method for stochastic optimization.
\newblock {\em arXiv preprint arXiv:1412.6980}, 2014.

\bibitem{kong2016photo}
Shu Kong, Xiaohui Shen, Zhe Lin, Radomir Mech, and Charless Fowlkes.
\newblock Photo aesthetics ranking network with attributes and content
  adaptation.
\newblock In {\em European Conference on Computer Vision}, pages 662--679.
  Springer, 2016.

\bibitem{lee2019image}
Jun-Tae Lee and Chang-Su Kim.
\newblock Image aesthetic assessment based on pairwise comparison a unified
  approach to score regression, binary classification, and personalization.
\newblock In {\em Proceedings of the IEEE/CVF International Conference on
  Computer Vision}, pages 1191--1200, 2019.

\bibitem{lee2020deep}
Seon-Ho Lee and Chang-Su Kim.
\newblock Deep repulsive clustering of ordered data based on order-identity
  decomposition.
\newblock In {\em International Conference on Learning Representations}, 2020.

\bibitem{levi2015age}
Gil Levi and Tal Hassner.
\newblock Age and gender classification using convolutional neural networks.
\newblock In {\em Proceedings of the IEEE Conference on Computer Vision and
  Pattern Recognition Workshops}, pages 34--42, 2015.

\bibitem{li2006ordinal}
Ling Li and Hsuan-Tien Lin.
\newblock Ordinal regression by extended binary classification.
\newblock {\em Advances in Neural Information Processing Systems}, 19, 2006.

\bibitem{li2021learning}
Wanhua Li, Xiaoke Huang, Jiwen Lu, Jianjiang Feng, and Jie Zhou.
\newblock Learning probabilistic ordinal embeddings for uncertainty-aware
  regression.
\newblock In {\em Proceedings of the IEEE/CVF Conference on Computer Vision and
  Pattern Recognition}, pages 13896--13905, 2021.

\bibitem{li2019bridgenet}
Wanhua Li, Jiwen Lu, Jianjiang Feng, Chunjing Xu, Jie Zhou, and Qi Tian.
\newblock {BridgeNet}: A continuity-aware probabilistic network for age
  estimation.
\newblock In {\em Proceedings of the IEEE/CVF Conference on Computer Vision and
  Pattern Recognition}, pages 1145--1154, 2019.

\bibitem{lim2019order}
Kyungsun Lim, Nyeong-Ho Shin, Young-Yoon Lee, and Chang-Su Kim.
\newblock Order learning and its application to age estimation.
\newblock In {\em International Conference on Learning Representations}, 2019.

\bibitem{liu2018ordinal}
Xiaofeng Liu, Yang Zou, Yuhang Song, Chao Yang, Jane You, and BV
  K~Vijaya~Kumar.
\newblock Ordinal regression with neuron stick-breaking for medical diagnosis.
\newblock In {\em Proceedings of the European Conference on Computer Vision
  (ECCV) Workshops}, pages 0--0, 2018.

\bibitem{liu2017deep}
Yanzhu Liu, Adams Wai-Kin Kong, and Chi~Keong Goh.
\newblock Deep ordinal regression based on data relationship for small
  datasets.
\newblock In {\em IJCAI}, pages 2372--2378, 2017.

\bibitem{liu2018constrained}
Yanzhu Liu, Adams Wai~Kin Kong, and Chi~Keong Goh.
\newblock A constrained deep neural network for ordinal regression.
\newblock In {\em Proceedings of the IEEE Conference on Computer Vision and
  Pattern Recognition}, pages 831--839, 2018.

\bibitem{liu2019probabilistic}
Yanzhu Liu, Fan Wang, and Adams Wai~Kin Kong.
\newblock Probabilistic deep ordinal regression based on {Gaussian} processes.
\newblock In {\em Proceedings of the IEEE/CVF International Conference on
  Computer Vision}, pages 5301--5309, 2019.

\bibitem{liu2021swin}
Ze Liu, Yutong Lin, Yue Cao, Han Hu, Yixuan Wei, Zheng Zhang, Stephen Lin, and
  Baining Guo.
\newblock {Swin transformer}: Hierarchical vision {Transformer} using shifted
  windows.
\newblock In {\em Proceedings of the IEEE/CVF International Conference on
  Computer Vision}, pages 10012--10022, 2021.

\bibitem{martin2014dating}
Paul Martin, Antoine Doucet, and Fr{\'e}d{\'e}ric Jurie.
\newblock Dating color images with ordinal classification.
\newblock In {\em Proceedings of International Conference on Multimedia
  Retrieval}, pages 447--450, 2014.

\bibitem{niu2016ordinal}
Zhenxing Niu, Mo Zhou, Le Wang, Xinbo Gao, and Gang Hua.
\newblock Ordinal regression with multiple output {CNN} for age estimation.
\newblock In {\em Proceedings of the IEEE Conference on Computer Vision and
  Pattern Recognition}, pages 4920--4928, 2016.

\bibitem{palermo2012dating}
Frank Palermo, James Hays, and Alexei~A Efros.
\newblock Dating historical color images.
\newblock In {\em European Conference on Computer Vision}, pages 499--512.
  Springer, 2012.

\bibitem{pan2019image}
Bowen Pan, Shangfei Wang, and Qisheng Jiang.
\newblock Image aesthetic assessment assisted by attributes through adversarial
  learning.
\newblock In {\em Proceedings of the AAAI Conference on Artificial
  Intelligence}, volume~33, pages 679--686, 2019.

\bibitem{pan2018mean}
Hongyu Pan, Hu Han, Shiguang Shan, and Xilin Chen.
\newblock Mean-variance loss for deep age estimation from a face.
\newblock In {\em Proceedings of the IEEE Conference on Computer Vision and
  Pattern Recognition}, pages 5285--5294, 2018.

\bibitem{radford2019language}
Alec Radford, Jeffrey Wu, Rewon Child, David Luan, Dario Amodei, Ilya
  Sutskever, et~al.
\newblock Language models are unsupervised multitask learners.
\newblock {\em OpenAI Blog}, 1(8):9, 2019.

\bibitem{raffel2020exploring}
Colin Raffel, Noam Shazeer, Adam Roberts, Katherine Lee, Sharan Narang, Michael
  Matena, Yanqi Zhou, Wei Li, Peter~J Liu, et~al.
\newblock Exploring the limits of transfer learning with a unified text-to-text
  transformer.
\newblock {\em J. Mach. Learn. Res.}, 21(140):1--67, 2020.

\bibitem{ramesh2021zero}
Aditya Ramesh, Mikhail Pavlov, Gabriel Goh, Scott Gray, Chelsea Voss, Alec
  Radford, Mark Chen, and Ilya Sutskever.
\newblock Zero-shot text-to-image generation.
\newblock In {\em International Conference on Machine Learning}, pages
  8821--8831. PMLR, 2021.

\bibitem{ratner2018learning}
Vadim Ratner, Yoel Shoshan, and Tal Kachman.
\newblock Learning multiple non-mutually-exclusive tasks for improved
  classification of inherently ordered labels.
\newblock {\em arXiv preprint arXiv:1805.11837}, 2018.

\bibitem{russakovsky2015imagenet}
Olga Russakovsky, Jia Deng, Hao Su, Jonathan Krause, Sanjeev Satheesh, Sean Ma,
  Zhiheng Huang, Andrej Karpathy, Aditya Khosla, Michael Bernstein, et~al.
\newblock Imagenet large scale visual recognition challenge.
\newblock {\em International Journal of Computer Vision}, 115(3):211--252,
  2015.

\bibitem{shin2022moving}
Nyeong-Ho Shin, Seon-Ho Lee, and Chang-Su Kim.
\newblock Moving window regression: A novel approach to ordinal regression.
\newblock In {\em Proceedings of the IEEE/CVF Conference on Computer Vision and
  Pattern Recognition}, pages 18760--18769, 2022.

\bibitem{simonyan2014very}
Karen Simonyan and Andrew Zisserman.
\newblock Very deep convolutional networks for large-scale image recognition.
\newblock {\em arXiv preprint arXiv:1409.1556}, 2014.

\bibitem{sutskever2014sequence}
Ilya Sutskever, Oriol Vinyals, and Quoc~V Le.
\newblock Sequence to sequence learning with neural networks.
\newblock {\em Advances in Neural Information Processing Systems}, 27, 2014.

\bibitem{vaswani2017attention}
Ashish Vaswani, Noam Shazeer, Niki Parmar, Jakob Uszkoreit, Llion Jones,
  Aidan~N Gomez, {\L}ukasz Kaiser, and Illia Polosukhin.
\newblock Attention is all you need.
\newblock {\em Advances in Neural Information Processing Systems}, 30, 2017.

\bibitem{wang2021pyramid}
Wenhai Wang, Enze Xie, Xiang Li, Deng-Ping Fan, Kaitao Song, Ding Liang, Tong
  Lu, Ping Luo, and Ling Shao.
\newblock Pyramid vision {Transformer}: A versatile backbone for dense
  prediction without convolutions.
\newblock In {\em Proceedings of the IEEE/CVF International Conference on
  Computer Vision}, pages 568--578, 2021.

\bibitem{wang2022pvt}
Wenhai Wang, Enze Xie, Xiang Li, Deng-Ping Fan, Kaitao Song, Ding Liang, Tong
  Lu, Ping Luo, and Ling Shao.
\newblock {PVT} v2: Improved baselines with pyramid vision {Transformer}.
\newblock {\em Computational Visual Media}, 8(3):415--424, 2022.

\bibitem{wen2020adaptive}
Xin Wen, Biying Li, Haiyun Guo, Zhiwei Liu, Guosheng Hu, Ming Tang, and Jinqiao
  Wang.
\newblock Adaptive variance based label distribution learning for facial age
  estimation.
\newblock In {\em European Conference on Computer Vision}, pages 379--395.
  Springer, 2020.

\bibitem{williams1976modification}
Louis~F Williams~Jr.
\newblock A modification to the half-interval search (binary search) method.
\newblock In {\em Proceedings of the 14th annual Southeast regional
  conference}, pages 95--101, 1976.

\end{thebibliography}
}

\end{document}